Taylor & Francis
Taylor & Francis Group

RESEARCH ARTICLE

OPEN ACCESS

# The Digital Transformation in Health: How AI Can Improve the Performance of Health Systems

África Periáñez, Ana Fernández Del Río, Ivan Nazarov, Enric Jané, Moiz Hassan, Aditya Rastogi, and Dexian Tang

Causal Foundry, Inc, Newark, DE, USA

**ABSTRACT**
Mobile health has the potential to revolutionize health care delivery and patient engagement. In this work, we discuss how integrating Artificial Intelligence into digital health applications focused on supply chain operation, patient management, and capacity building, among other use cases, can improve the health system and public health performance. We present the Causal Foundry Artificial Intelligence and Reinforcement Learning platform, which allows the delivery of adaptive interventions whose impact can be optimized through experimentation and real-time monitoring. The system can integrate multiple data sources and digital health applications. The flexibility of this platform to connect to various mobile health applications and digital devices, and to send personalized recommendations based on past data and predictions, can significantly improve the impact of digital tools on health system outcomes. The potential for resource-poor settings, where the impact of this approach on health outcomes could be decisive, is discussed. This framework is similarly applicable to improving efficiency in health systems where scarcity is not an issue.

## Introduction

Low- and middle-income countries (LMICs) face multiple challenges in health systems and public health, mostly caused by constrained resources, inadequate infrastructure, and socioeconomic inequalities. One major issue is poor access to basic health care services, often resulting from a scarcity of health care facilities and trained health care professionals.[1] Ineffective health information systems in LMIC hamper the effective collection and utilization of data needed for decision-making and resource allocation.[2] Health disparities are aggravated by socioeconomic factors, resulting in the poorest populations having the least access to essential health care services and increased disease burdens.[3] Infectious diseases, including HIV/AIDS, tuberculosis, and malaria, disproportionately affect LMICs, straining their health systems and diverting scarce resources from other important public health initiatives.[4] Furthermore, the rise of non-communicable diseases is creating a dual burden, complicating efforts to enhance public health in already-constrained circumstances.[5] High-Income Countries (HICs) also face significant health care challenges. Pervasive issues related to patient engagement, adherence to treatments, and adoption of public health interventions prevail.[6]

Accelerated by the COVID-19 pandemic, the global health care sector is experiencing rapid growth in the use of digital health tools as promising solutions to many persistent challenges encountered by patients, caregivers, and suppliers.[7] As part of the impetus toward Universal Health Coverage (UHC), the World Health Organization has recognized a set of prioritized digital interventions for health system strengthening, including health worker decision support and targeted provider and patient communication via mobile applications.[8] These platforms can support various health system functions, such as communication, management of patients and chronic conditions, self-reporting, adherence support, access to medical knowledge bases and pharmaceutical catalogs, capacity building, procurement of drugs and medical supplies, and inventory and distribution management.

In recent years, Artificial Intelligence (AI) has emerged as a powerful tool in health care, capable of utilizing data for decision-making and personalized interventions.[9–12] AI systems can use adequately tracked data from digital health platforms to offer insights, predictions, and recommendations into patient and provider behaviors, health care quality, and demand, facilitating the development of adaptive interventions delivered through digital tools.[13–17] AI-enhanced adaptive interventions are not only tailored—they also continuously evolve, adapting in real-time to each user's unique choices and circumstances. This evolution determines the content and timing of interventions and enables the identification of individuals who may need additional support. Furthermore, AI can be used to anticipate medication demand variations to help in inventory management. Enabling prediction-based timely reminders and guidance systems can promote consistent availability of vital supplies across all distribution channels.[18]

AI can also be used to harmonically combine and reconcile information from multiple sources, including various digital tools that contain rich information on the health care system's status, functioning, and efficiency at different levels. Merging these and other sources of information into a holistic picture can facilitate evidence-based decision-making, generating a framework that policymakers can use to systematically explore, understand, and quantify the real or potential impact of different current or planned public health strategies and health system policies. AI technologies can also be used in experimental designs for robust risk-minimizing public policy evaluation. To date, digital adaptive strategies are predominantly used in HICs, particularly in entertainment and e-commerce.[19,20] However, the expanding mobile health infrastructure and surging smartphone adoption, including in LMICs, present a significant potential for positive global health impacts through adaptive measures. Indeed, an increasing number of digital and adaptive intervention initiatives have been appearing in and targeting the Global South in recent years.[21-25]

This article presents an AI platform designed to generate personalized predictions, recommendations, and insights by harnessing and analyzing data from digital health tools and other relevant sources. The platform provides the capability to create, test, and deploy adaptive interventions driven by Reinforcement Learning to enable optimized user engagement, resource allocation recommendations, and clinical and behavioral guidance for users. These capabilities are made possible through the platform's integration with existing digital tools designed to support patients, medical facilities, pharmacists, Community Health Workers (CHWs), midwives, and distributors of drugs and medical consumables by enhancing their operational efficacy. This paper is intended for a public health audience, focusing on the practical benefits and outcomes relevant to the public health context. While references are provided for those interested, it does not delve into the technicalities and specificities of the AI and Machine Learning (ML) algorithms and methods utilized by the platform.

While the discussion in this paper is framed specifically around the potential impact on health systems in LMICs, the same approach would also benefit health systems in HICs by driving efficiency and better health outcomes. While the use of AI for certain use cases is more widespread in these contexts, an integrated approach that would unleash the full potential of this framework is yet not in place for even the most well-resourced health systems.

## Digital Health and Artificial Intelligence

AI and ML are becoming essential everywhere, including increasingly being used in public health and health care. AI has recently received significant attention due to the popularization of large language models (LLMs). At their core, AI and ML are techniques that enable systems to learn patterns from large sets of data without being explicitly programmed for specific tasks. In most cases, these tools are designed to make predictions from large amounts of data without focusing on understanding the underlying mechanisms driving the system outputs.[26] Several domains within AI have been developed to address different problems, such as natural language processing, time series forecasting, and computer vision. AI is a rapidly evolving field with an active research community constantly exploring new applications underpinned by increasing computational power. Given the large amount of data generated by digitized health systems, harnessing AI to extract meaningful insights and make evidence-based decisions has become a major frontier in public health and health care.[27-30] Potential applications specifically for global health have gathered attention in recent years.[9,12]

Predictive analytics is an important ML component that allows public health professionals to forecast disease and potential outbreak trends. For instance, ML algorithms can be trained on historical epidemiological data to predict future disease hotspots or identify factors that make populations susceptible to health challenges. This forward-looking capability could be used to support optimal resource allocation and timely deployment of public health measures to protect vulnerable populations more efficiently with targeted interventions.[31-33]

Another transformative application lies in behavioral nudges and other personalized interventions.[34-37] By analyzing individuals' behaviors, habits, and preferences, ML algorithms can craft personalized messages or recommendations designed to encourage the adoption of best practices, healthier choices, and adherence to treatment and public health interventions, among other behaviors.[14,38] These systems dynamically adapt, allowing rapid innovation cycles to support the deployment of agile and responsive public health and health care systems. Figure 1 illustrates how an adaptive intervention could improve an undesired health care outcome.

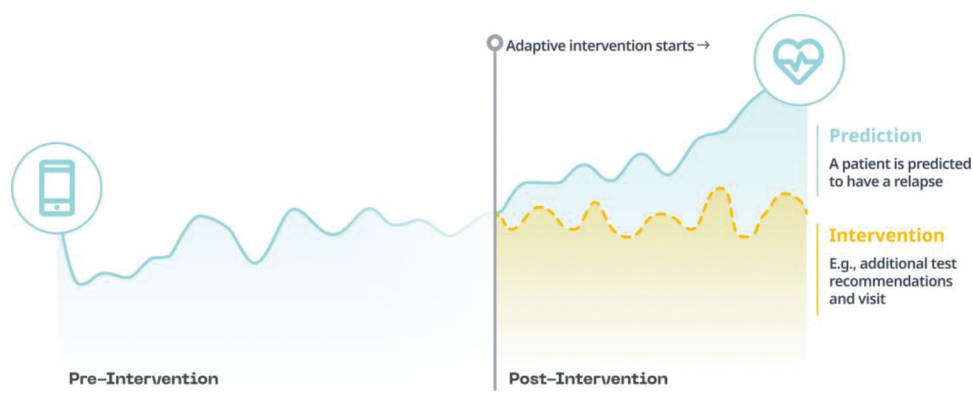

Figure 1. Adaptive intervention. The goal of adaptive interventions is to dynamically adjust the strategies used to individuals' changing needs and context over time. The adaptations are selected using data-driven machine learning methods that analyze previous responses to interventions and predicted and observed performance to maximize overall performance.

Finally, AI can play a role in addressing the challenges of digital health fragmentation, where systems often operate in silos with limited interoperability. For example, CHW job aids, procurement and supply chain management apps, electronic health records, and customer-facing tools serving patients and caregivers all generate vast data. AI systems are capable of integrating diverse data streams and then offering contextualized insights. When exploited by AI, these data can significantly enhance the effectiveness and efficiency of health systems. The following subsections expand on how AI could be leveraged to improve the health system's efficiency and outcomes.

### Primary Care Facilities and Health Worker Engagement and Support

Strong community-based integrated primary health care (PHC) is one of the cornerstones of UHC, quality of care, and improved health outcomes. Yet in LMICs, primary health care facilities often lack the necessary human and material resources to operate adequately.[39] Health care workers (HWs), especially in LMICs, grapple with many challenges, including fragmented programming, lack of continuous performance assessment, and the overwhelming task of managing diverse patient needs with limited resources. In addition, many LMIC health systems struggle to ensure that HWs are trained, supervised, and equipped to provide quality care.

Digital tools can support HWs and PHC facilities by offering guidance on clinical processes and connecting them to patients and peers. They can also be used for continuous capacity building and professional development, operational tasks (such as appointment management and referrals), the provision of required tests and medicines, and stock management. Use of patient management and clinical decision support tools can also facilitate reporting and promote adherence to established clinical guidelines and efficient resource allocation.

Examples of tools that are already being used by HWs and facilities are: CHARM[40] (Client & Community Health Assisted Resources for Mentors), a job aid used by Mothers2mothers' mentor mothers in Angola, Ghana, Kenya, Lesotho, Malawi, Mozambique, South Africa, Tanzania, Uganda and Zambia; Lifebank's Nerve App,[41] which allows hospitals and clinics in Nigeria to conveniently order blood, oxygen and medical consumables online); Ilara Health's HMIS[42] (Health Management Information System), a digital tool for streamlining clinic operations in Kenya via guided consultations, electronic health records, patient billing and inventory management; Global Strategies' NoviGuide,[43] a neonatal care clinical decision support software available worldwide; and Maternity Foundation's Safe Delivery App,[44–46] which is a capacity building and job aid for skilled birth attendants, available worldwide with a generic version with additional language and context-specific versions for multiple countries.

Integrating AI and ML into the digital tools used by HWs offers promising avenues to improve their ability to provide quality health care.[22] Personalized HW interventions can support adaptive learning journeys, timely reminders for crucial tasks, incentives for best practices, and patient-specific recommendations based on clinical and behavioral data. For example, AI-powered digital tools can use patient data to assist HWs in identifying high-risk patients who could benefit from extra support and targeted communications. This is particularly important in the case of CHWs, who play a crucial role in linking local, and often, remote communities with the formal health care system, and who usually have limited tools and skills.

When it comes to improving the skills of HWs, one challenge lies in ensuring their adherence to digitally enabled training programs. ML tools allow for the building of adaptive learning journeys, in which user data is utilized to design individualized nudges and content to ensure they are engaged and able to pass every certification.

Finally, AI systems can support HWs by offering patient-specific recommendations, including regarding clinical workflows (e.g., suggesting additional testing for diagnosis or before prescription or referral). While this paper focuses on operational and behavioral aspects of health care and public health, the AI functionalities presented in the following sections are equally applicable for purely clinical interventions.

### Patient and Caregiver Engagement and Support

Patients and caregivers in LMICs face daunting challenges when seeking life-saving health care, including timely access to commodities and medical advice. Digital apps can support them by connecting them to relevant health information, health care providers, and pharmacies.

In this case as well, the digital tools generate vast amounts of potentially useful data on patients' health, the information they seek, and other behaviors. Take, for example, the potential contributions of AI to improving family planning programs. With increasing access to mobile technology in these countries, particularly among the youth, AI-driven apps could provide teenagers and young adults with tailored advice on contraceptive choices, potential side effects, and reminders on proper usage. When the apps are integrated with pharmacy chains' digital systems, they could also facilitate access to contraceptives, decreasing barriers related to social norms.

There are already examples of deployed and widespread solutions in this arena, such as: Momentum,[47] used in Ghana to order drugs online from the Aide Chemist chain; Vitala Global's Aya Contigo,[48] a virtual sexual and reproductive health app available in Venezuela and the USA; Chekkit App,[49] an anti-counterfeiting solution for medicines and other products; and, Appy Saude,[50] an app- and web-based tool for online ordering of medicines and other products available at a network of pharmacies in Angola.

By centering patients' needs when offering AI-driven solutions to address challenges, the health care landscape in LMICs can be significantly transformed, ensuring that patients receive robust support.

### Pharmacies

Pharmacies are often the first line of contact for patients seeking health care. Pharmacists are key HWs, ensuring patients and caregivers get proper medical advice, adhere to treatment, and responsibly use medicines. In many LMICs, one challenge is ensuring pharmacies stay abreast of treatment guidelines and offer quality care to patients. Digital tools again offer opportunities to provide training and to gather data on pharmacy customers. AI can leverage these data to ensure pharmacists receive nudges and incentives to follow guidelines and provide quality care, strengthening pharmacies' role in the overall health system.

Pharmacists are business operators as well as HWs, and often struggle with inventory management, forecasting, and procurement optimization, among other critical functions. These activities involve proactive collaboration with pharmaceutical warehouses in order to replenish stock while judiciously balancing demand against the risk of overstocking, which can lead to wastage through expired medications. AI systems can support all these functions, ensuring pharmacies are well-stocked with life-saving commodities.

A noteworthy trend in LMICs is the growth of tech companies specializing in business-to-business (B2B) e-commerce distribution to avoid dependence on expensive intermediaries to acquire supplies. Simultaneously, many pharmacies are evolving into franchises and chains. E-commerce applications for health care products are becoming a convenient (and often cheaper) way to purchase and track goods. AI can leverage the data on demand and stock movement generated through these digital tools to help pharmacists manage stocks to ensure essential medicines are always available.

Notable examples of such platforms are SwipeRx (a comprehensive tool for Southeast Asian pharmacies that includes B2B e-commerce, professional networking, continuous professional development, and more) and Field's Shelf Life (a supply service for pharmacies in Kenya and Nigeria operating both on-demand and by subscription models).[51,52]

### Supply Chains

Supply chains are a vital component of health systems. Efficient supply chains prevent wastage and counterfeiting, safeguard against stockouts, and ensure that life-saving commodities, such as antimalarials and antibiotics, are available at the point of care. However, the intricacy of these networks, especially in LMICs, often results in stockouts across the supply chain and at the point of care.[53] The prevalence of substandard and falsified drugs is also an issue in some markets.

Digital tools have significant potential to make the supply chain more efficient. They can ease procurement through B2B solutions and reduce paperwork, using technologies such as computer vision or radio frequency identification (RFID) to improve inventory management and prevent the circulation of substandard and falsified medicines. Software for supply chain processes facilitates the storage of reliable data with integrity, enabling its systematic use for knowledge extraction and intervention. Field's Supply is an example of such a supply chain software,[52] as is Chekkit, a blockchain-based solution for consumers to verify the authenticity of drugs and other products.[49]

AI can integrate real-time data from multiple digital applications, including inventory levels, demand patterns, and other factors like climate events or disease outbreaks, to understand their evolution and predict future needs. Adaptive interventions could promote timely medication replenishment and aid in detecting abnormal behaviors or potential supply disruptions early. This type of model empowers stakeholders to make informed decisions to optimize the supply chain from procurement to delivery.[54]

AI-powered apps can generate tailored support for customers, distributors, and other key actors in the supply chain system, including reminders, product recommendations and prioritization, and order and distribution plan proposals. These interventions can leverage demand and user behavior predictions to make the adaptive framework optimize distribution operations and reduce stockouts of essential medications at points of care.

### Policymakers and Health Authorities

The previous sections have highlighted the effects that digital tools can have on the day-to-day work and efficiency of health care delivery, focusing on the potential of AI to magnify and scale their impact. This is bound to have a noticeable effect on the health system's performance as a whole. However, we also expect that there will be more benefits from the use of these technologies. AI can be used to organize and learn from many sources of information readily available to health authorities and other health systems decision-makers, such as public facility periodic reporting and performance-based financing. A combination of incentives and regulations can be used to promote the integration of additional sources of other relevant information. All these sources can be combined and transformed to provide relevant insights and a bird's eye view of the health system's functioning to health authorities and other stakeholders, such as health insurance or international cooperation agencies.

Efficiency, performance, and outcome indicators can be defined and monitored at different levels (practitioner, facility, and regional). A system of alerts can be built through outlier detection to direct decision-makers' attention to issues needing intervention and can provide examples of successful interventions to consider. Data can be used to better understand the specific needs of a population, the availability of material and human resources, and the challenges, strengths, and financial health of different facilities or systems. This knowledge can then be used to more efficiently allocate resources (including human, medicines and medical equipment, funding, training, and capacity building) and to design strategies to tackle specific issues (e.g., inequities among facilities or neighboring regions in stockout frequency of an essential medicine, of referrals, or of neonatal mortality).

Adding a predictive layer on top of this integrated information system further enhances the framework's potential for aiding in planning, including enabling simulations of the effects of different policies and decisions under various scenarios. The predicted impact of any policy change can be further validated through experimentation with adaptive designs, allowing for safer rollouts.

It is beyond the scope of this paper to discuss in detail which specific indicators are the best ones to assess and to track health system performance and guide policy-making or reform. Different contexts will require different levels of attention to different types of indicators. In general, however, they should include monitoring of:

- Accessibility and coverage (e.g., What fraction of the population is insured? Are out-of-pocket expenses affordable? How far away is the nearest health facility?)
- Quality of care (e.g., Are clinical guidelines being followed? Is the level of staffing reasonable? Are HWs adequately trained? Are essential medicines and tests available and affordable?)
- Efficiency and sustainability (e.g., Are cases treated at the appropriate level of care? Is care provided at the lowest possible cost? Are there prevention efforts in place? Are financing mechanisms reliable and predictable?)

Attention should be paid both to average values and to variation across regions, facilities, and population groups.

### A Reinforcement Learning Platform for Digital Health

Given its enormous potential to improve health systems, our company Causal Foundry proposes an AI platform designed to support health systems (including health care facilities, HWs, pharmacies, patients, and supply chains) in LMICs. This Reinforcement Learning Platform is the product of our efforts to ensure that the potential of AI can be easily integrated into the growing number of digital health tools in order to achieve the benefits described in the previous section. Our platform is designed to integrate with existing software, which reduces the time needed to deploy and scale effective interventions and provides a bridge connecting the different specialized systems that often contribute to the fragmentation of health systems. The result is a platform that allows the use of ML to increase the effectiveness and efficiency of the health system and improve patient outcomes.

In addition to predictive modeling, one major feature of the platform is that it allows testing and deploying adaptive interventions using digital tools. By adaptive, we mean that the system continuously tailors the intervention content and timing to each user based on the feedback collected from the ensemble of users. These interventions include sending messages (SMS, WhatsApp, push notification, in-app messages, etc.) to its users to promote behaviors such as adherence to treatment, reading or listening to relevant health information, or taking a training course.

As portrayed in Figure 2, the platform assists users to integrate data from different sources, use ML tools to get insights from the data, and create, test, and deploy adaptive interventions using Reinforcement Learning (RL) techniques. These interventions are pushed to the digital tools connected to the platform, and data collected prospectively then feeds back into the system, allowing it to learn and iteratively adapt the interventions to maximize the intended outcome. We provide an overview of the Causal Foundry platform's capabilities in the following subsections and conclude with a description of the current development and deployment status.

## The Platform

The primary aim of our platform is to integrate vast data sets (including maximizing their quality, proper format, and label) from digital tools (including behavioral, clinical, and contextual data) in order to provide valuable insights to stakeholders and create, test, and deploy personalized interventions (such as nudges, recommendations, and in-app personalized content). The platform ensures that acquired data are well organized and ready for use by ML predictive models. This emphasis on data collection and the use of opinionated data schemas is at the core of the platform functionality.

The digital platform is built on three foundational components (see Figure 3):

- **The user-facing frontend** provides a comprehensive user interface, with data and results visualization
- **The backend** serves as the operational core (depicted as the pipelines, database, and algorithms within the gray box in Figure 3)
- **The software development kit (SDK)** is the crucial bridge between the digital tools (end user in the Figure) and the platform

Each component is essential to ensuring users can seamlessly access, analyze, and manage their data and deploy ML models and real-time interventions. Below, we provide additional details for each of these three components.

The *frontend* has an intuitive user interface for data analysis and model and intervention management and monitoring. Users can perform behavioral and clinical data analysis, statistical modeling, and create cohorts of users. Furthermore, the interface enables designing, testing, and deploying interventions, such as nudges and rewards. It guides the user from a subject cohort (HWs, facilities, pharmacists, etc.) through how to analyze past behavior, predict future or target interventions, or define samples through algorithm selection to feature selection and target specification, including nudge alternatives in the case of interventions. The platform dashboard presents various traits, predictions, and metrics in an intuitive format to allow clear interpretation of the predicted results and impact of interventions.

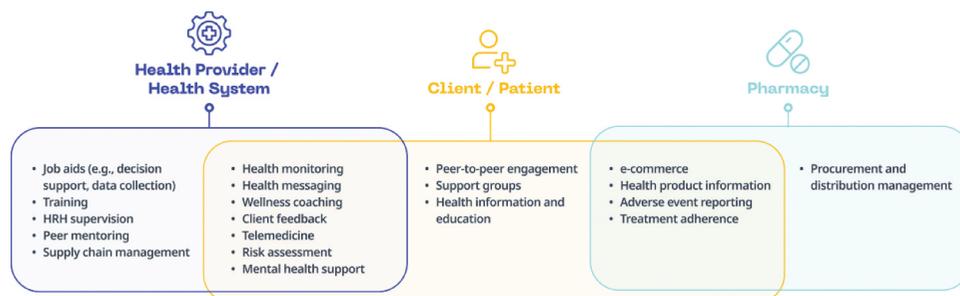

**Figure 2.** Examples of use cases that can benefit from the reinforcement learning platform. The RL platform delivers AI-based recommendations and incentives to assist health systems stakeholders, including patients, health care providers (e.g., frontline clinicians, nurses, midwives, community health workers, and pharmacists), pharmacies, and suppliers. It optimizes user engagement and resource use, leveraging data generated by an ecosystem of applications such as the ones shown.

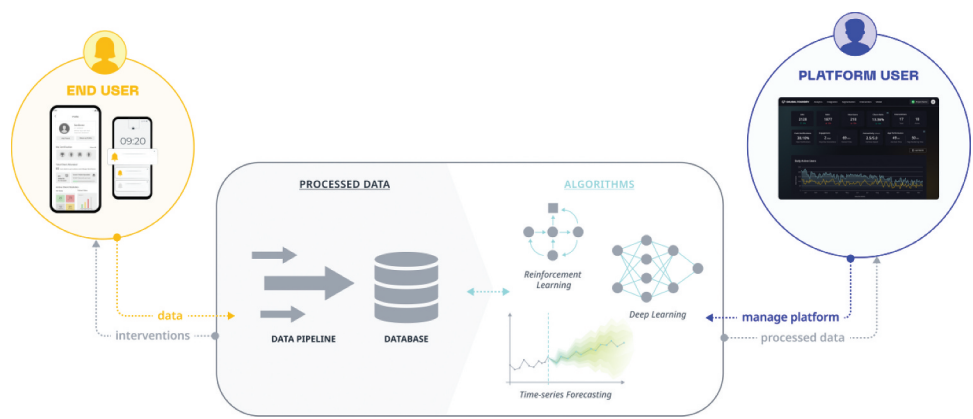

**Figure 3.** The reinforcement learning platform structure. The platform creates a robust data structure that logs user events on digital health devices (e.g., mobile apps), classifying them into dynamic and static traits. The backend system analyzes and processes these traits for statistical learning and machine-learning predictions and to generate adaptive interventions based on RL algorithms. User reactions to interventions are recorded via digital health devices and tracked through the system to feed back into the RL algorithms.

The *backend* is the platform's core, allowing the organization of data and the implementation of ML models and adaptive interventions. Data are collected and organized using domain-specific data schemas optimized for ML model implementation. Data security and privacy are key aspects captured in the backend. The platform's standardized design of event logging enables it to be a uniform interface for transformation and aggregation across different application domains (e.g., supply chain and e-commerce marketplace, medication tracking, and patient communication). The backend also hosts the predictive modeling engine that handles the ML component and the algorithmic decision-making service. These modules configure, train, host, and manage models for time series forecasting, deep and ensemble survival analysis, and sequential decisions via reinforcement learning.

Finally, an open-source *SDK* tracks the data with the correct labels and formats, creating a data pipeline between the digital tools and the platform.[a] This SDK is a lightweight software embedded into mobile apps, so data are tracked directly from the source. The software is optimized for the variety of mobile devices used in LMICs, such as budget smartphones and tablets, minimizing the impact of platform integration on device performance. This SDK provides tools for collecting relevant events and messaging services that deliver the interventions. The tool does not collect Personal Identifiable Information (PII) and complies with General Data Protection Regulation (GDPR) and the Health Insurance Portability and Accountability Act (HIPAA) guidelines. It tracks user actions as time-series data and contextual information such as details about the medicines prescribed, viewed, or purchased. For interventions, the SDK is responsible for presenting the nudges to the mobile app user and automatically logging their reactions (e.g., if they opened, viewed, discarded, or blocked the nudge), ensuring complete visibility of the intervention lifecycle.

The SDK uses a functionality-specific opinionated data model for data collection. This model is optimized for data science and ML purposes and ensures stringent control over content and quality. The goal is to ensure that: everything needed to tackle the relevant use cases for each functionality is recorded in the required format; no redundant or unnecessary information is tracked, in order to function without compromising app performance, even with limited connectivity, and to reduce storage needs; and, most importantly, the operational data pipelines are as simple and less prone to error as possible. The SDK is also responsible for maintaining interaction logs in offline settings and autonomously managing the network layer to ensure logs are synchronized with the Causal Foundry platform as soon as the device regains connectivity. This functionality alleviates the need for app developers to implement networking capabilities for offline usage and supports HWs by eliminating the necessity of manually opening the app to sync data. The modular (i.e., with different components for different functionalities) nature and generality of the SDK allow for integration with a wide variety of existing tools, while the rigidity of its data model enables comparability across applications and cross-compatibility of any models and interventions developed.

### *Predictive Modeling*

Predictive modeling is crucial for defining who should be targeted with specific interventions and optimizing their timing. The Causal Foundry platform allows users to utilize three prediction models: time-to-event prediction (also known as survival analysis), demand forecasting, and recommendation systems, such as click-through-rate predictions or causal graphic analysis.[b]

*Time-to-event prediction* offers a way of characterizing behaviors by modeling the probability of occurrence of different events of interest and the duration between them.[34,55–57] For example, it can be used to predict the risk of non-adherence to treatment and the time when it will happen, client churn rate in an e-commerce platform, or a health worker failing to complete a training module. These types of predictions can be used to identify target subjects for specific interventions (e.g., reminders to support treatment adherence, promotions to discourage churn, motivational prompts, or incentives to encourage learning).

The second model class, *demand forecasting*, is crucial in e-commerce, supply chain, and inventory optimization. It includes, for example, multivariate forecasting methods that learn either point or probabilistic forecasts from codependent time series to predict demand for different pharmaceutical products.[58] Several approaches are available through our platform, including autoregression and graph-regularized factor analysis and latent state forecasting, state space modeling with long short-term memory (LSTM) and Gaussian copula for probabilistic forecasts, recurrent networks for multivariate count data with non-uniform scale, and attention-based deep models for multi-horizon interval predictions.[59–63] Observed stocks and previous ordering behavior, together with the predicted evolution of demand, can be used to identify which pharmacies need reminders to restock or can suggest products to add to the cart while ordering is underway. Similar approaches can be used for patients using e-commerce services and to support decisions on distributing stocks across warehouses.

Finally, *click-through-rate predictions* allow the platform to generate traits useful for content personalization. Mixed deep learning-based and factorization methods have shown good performance,[64–67] In the next section, we discuss how adaptive delivery will be the main mechanism to learn and exploit user preferences for recommendations and personalization. Other recommendation algorithms are also available in the platform, as they can provide valuable insights to be leveraged by the RL algorithm. Other models, such as *causal graph analysis*, can also help to understand cause-and-effect relationships. For example, which factors caused an event? Why did a particular user stop using the app? Which patient is more likely to stop a treatment? Known causes can drive the design of personalized interventions to reduce the chance of that event occurring in the future.

### Adaptive Intervention Delivery and Experimentation

One of our platform's key features is the ability to design, test, and deploy adaptive interventions. Adaptive intervention delivery uses individual and contextual data for effective nudging and interventions. Let us illustrate the main underlying ideas using a concrete example of a digital tool, such as an app, for pregnant women and mothers through which they can report information on their pregnancy and newborn (e.g., overall well-being, health issues, vaccinations, use of health care services) and receive relevant health information. The tool allows us to detect any health issue the women and their newborns might have and provide them with relevant health advice (such as recommendations on when to seek care). The impact of this patient engagement tool hinges on women using the tool—however, we know user engagement is challenging to maintain.[68] Through formative research, behavioral science analysis, and focus group discussions, we decide that we want to test two interventions, one that nudges women with recommendations related to their characteristics (stage of pregnancy or age of their newborn), and another that sends messages with suggestions based on behaviors exhibited by similar users.

We then need to determine how to measure the intervention's success, and decide to use the information women submit through the tool. Finally, we need to select the characteristics (collected through the digital tool) that could influence how impactful a particular nudge will be at any given time. These could include characteristics of the women and their babies (e.g., gestational age, number of previous pregnancies) and their use of the digital tool (e.g., recent exposure and interaction with nudges, information they submitted to the app, and which content they never use). We refer to these characteristics as contextual traits.

The adaptive intervention delivery then goes into action. Each week (or as scheduled when defining the intervention), the platform decides which nudges to send to each user based on the contextual traits and observes how users with different characteristics react (based on how we decided to measure success) in various situations (as defined by the contextual traits). The platform algorithm will continue to refine and exploit information on what works best in any given situation. It will also detect changes in how users react to the nudges and adapt correspondingly.

At the end of the experiment, we will be able to select the interventions that were most successful in increasing engagement, and discard those that did not generate a significant change. The selected intervention(s) will then continue to learn over time as the platform collects additional feedback. This example illustrates how the platform's intervention and experimentation capabilities can be used to assess the impacts of different intervention strategies.

The platform can use a sample of users to perform rapid cycles of experimentation for intervention refinement and subsequently roll out the final intervention to all users. Digital trials can be easily designed and conducted through the platform, including near real-time monitoring, with impact estimation and heterogeneous effects analysis. The platform includes both fully randomized and adaptive trial designs. Notably, adaptive experiments do not need to stop. In randomized trials, such as A/B testing, the researcher pauses the experiment to measure which treatment is statistically significantly better than others, and decides accordingly. In adaptive studies like the one described above, it is possible to run them continuously, making them select the best-responding treatment. The algorithm automatically selects the most conditionally effective treatment and adapts to the non-stationarity of the treatment response.

Adaptive designs also maximize statistical power and minimize risk, as the algorithm considers experimental outcomes already observed and reduces the number of subjects assigned to treatment in the event of an adverse effect on the success measure. Experimental designs that assign each subject multiple times throughout the experiment can be better options when studying the immediate impact of repeated interventions (e.g., messages sent with a specific frequency), while trials with a single assignment and a control group remain the golden standard to understand one-shot interventions and long-term effects.[69–76]

Figure 4 shows a monitoring plot for an experiment conducted on health product recommendations delivered to pharmacies via an app for ordering medicines from a distributor. The recommendations aimed to increase the weekly variety of products ordered by pharmacists. They recommended pairs of items that are typically ordered together when the pharmacy receiving the recommendation only purchased one of them regularly. The plot shows the daily mean difference in the weekly purchased variety (the number of different health products ordered in the previous seven days), and the shaded confidence intervals are related to hypothesis testing to establish differences in means. The bar plots show the number of messages interacted with daily. The sample sizes in the trial were small, but the intervention had a significant (95%) positive impact during the first two weeks. Pharmacies receiving the recommendations purchased, on average, five more different health products than those that did not receive recommendations. The trial also showed indications of a minor positive effect later (i.e., the impact of the recommendation lasted days and even weeks after receiving the message) and fatigue (i.e., decreasing impact of the recommendations as the novelty disappears, leading eventually to adverse effects if the frequency of notifications is too high). All these findings were corroborated in successive rounds of experiments.

The Causal Foundry platform thus offers significant flexibility in designing experiments. For example, for interventions with facility HWs, in which interaction between participants at the same site is expected to create cross-contamination, the platform allows for cluster randomization (i.e., assignment to either treatment or control to all participants of the same facility). It also allows pairwise matching, an assignment mechanism typically used to achieve sample balance, notably in combination with cluster randomization, as it alleviates the need to increase sample sizes to achieve the same level of statistical power due to intracluster correlation. In this case, the underlying idea is to select pairs of facilities that share characteristics that we hypothesize will impact the outcome of interest, and then assign one to the control and one to the treatment arm for comparison.[76–78]

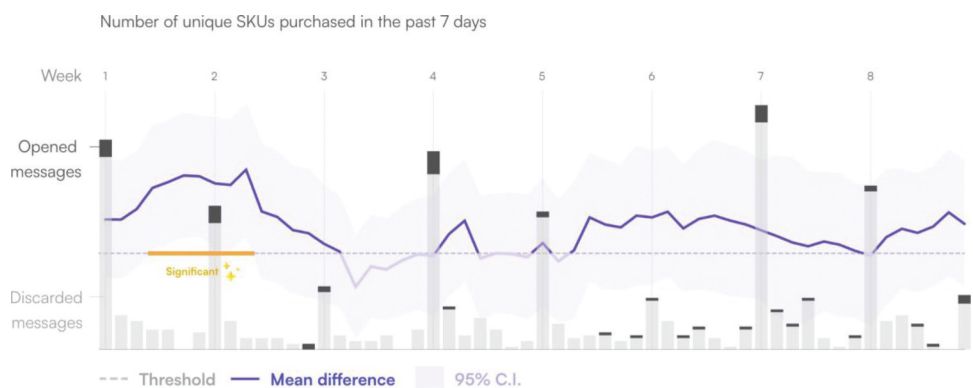

**Figure 4.** Improving the variety of products in stock. This figure presents results from a platform experiment in which the goal was to send recommendations to pharmacists to increase the variety of products in stock (known as SKUs). The daily mean difference between the weekly purchased variety of pharmacies in the treatment group (receiving periodic in-app messages with item recommendations) and in the control group (no recommendations received) is plotted as a line, together with the 95% confidence intervals. Bar plots indicate the number of users who interacted with the messages daily. The intervention was significant over the first two weeks. On average, those in the treatment group purchased five more unique items in the past week than the control group, including items never purchased before.

### Bandit Algorithms

Adaptive interventions are typically modeled as a Markov Decision Process (MDP), which makes RL the appropriate algorithmic paradigm.[79] RL-based interventions balance between-outcome optimization and knowledge extraction and adapt to the feedback and observed covariates of the intervention subjects. Within RL, *bandit algorithms* are particularly well suited for the problems described in the previous section. Bandit algorithms can be thought of as *online optimization methods with built-in model identification from partial feedback*. That is, the algorithm makes adaptive sequential decisions (based on the past observed covariates and interaction history) to eliminate suboptimal choices.[80]

On our platform, the algorithm analyzes the available features (static and dynamic traits) of each observed subject, decides on the allocation or type of intervention, and later receives a feedback score on its prior decision, which it uses to adapt and improve its decision mechanism. Possible applications of adaptive algorithms include sending reminders about treatment adherence and then scaling them back when a notification has been sent too recently or is not generating interaction. The analysis of bandit algorithms is beyond the scope of this paper, but there are plenty of resources exploring these algorithms, their use, challenges, and solutions.[80–106]

### Current Development Status

Causal Foundry's AI platform described in this paper is, as of June 2024, fully operational, integrated with up to ten digital tools, and growing. Its open-source SDK includes modules to track data from patient management, capacity building, e-commerce, supply chain, and condition management (patient self-tracking) tools. The core module covers activities common across all types of applications. The module- (and, to a large extent, use-case-) specific data pipelines create hundreds of additional indicators at the user, HW, facility, patient, drug, or medical supply level that can be visualized in the platform and used for cohort creation, as features for ML models and for intervention.

Four classes of ML models can be easily defined, trained, deployed, and monitored using the platform: time-to-event prediction algorithms, neural embedding-based item recommendation, contextual bandit algorithms for adaptive personalization and experimentation, and restless bandit algorithms for optimal and equitable resource allocation. The platform's adaptive intervention capabilities have been tested with multiple interventions, many of which have had a significant positive impact, and several other tests are planned. The platform also includes an Assistant that uses the generative models described in Section 4. The Assistant allows platform users to conversationally query all the information available through visualization throughout the platform. It can provide, for example, replies in English to questions such as: What were the ten facilities with a higher patient-to-practitioner ratio last week in Lagos? Or how many units of antimalarials are available at pharmacy X?

The platform is under constant improvement based on use cases. Causal Foundry works closely with partners to understand their greatest challenges and identify where there is potential for impact of AI technologies. The prioritization of new functionalities is also guided by the potential to generalize across use cases, tools, and contexts. Over the next months and years, we will deploy additional RL-based intervention capabilities, statistical tools for intervention and observation analysis, predictive and recommendation models, and visualizations in order to facilitate meaningful exploration of all available information.

### Generative AI

*Generative AI* (GAI) technologies use deep learning models trained on massive amounts of data to generate new text and images. *Large Language Models (LLMs)* are part of GAI and are trained on large datasets consisting of text. Once trained, GAI models can generate coherent, contextually relevant, and grammatically correct texts, answer questions, and perform text-based tasks. LLMs can revolutionize how HWs interact with information systems, enabling more efficient access to medical data and supporting decision-making. Integrating LLMs in HWs' mobile applications can leverage the benefits of a chat-based interface to facilitate user interaction and access to information more easily. Chatbots are integrated into the Causal Foundry AI platform to employ LLM-based functionalities, as depicted in Figure 5.

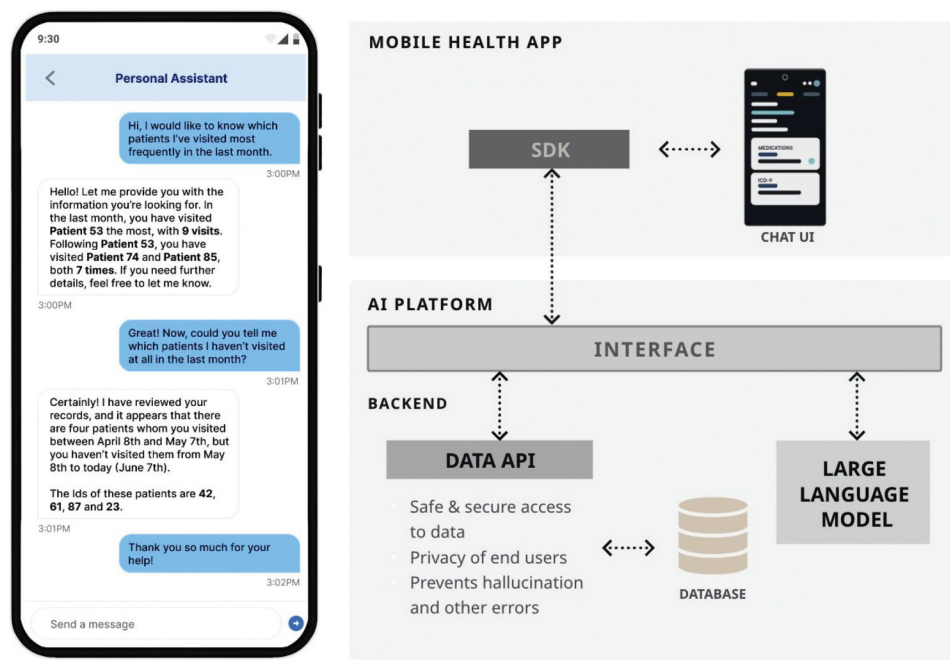

**Figure 5.** Integration of LLMs into the platform. The image on the left is an example of a health care worker interacting via an app chatbot with a Large Language Model (LLM). It shows how users can engage with the LLM to obtain relevant information or perform specific tasks, like creating an intervention. We can see that the LLM can handle each scenario, including understanding user intent, generating queries to the database, and returning the requested information. The diagram on the right is the proposed system architecture.

This product aims to empower CHWs, midwives, pharmacists, and clinicians by providing insights from data related to their practices or patients. The LLMs are pre-trained with internal data and predictions, based on mobile application usage, that are stored and organized by the AI platform. HWs can access the chatbot embedded in their digital tools to obtain answers to questions about the past or future in order to help with supply ordering and patient management. Examples of such questions include: Which patients have I not visited in over three months? Which patients did I refer in the past two months? How many HIV tests should I order for the next four weeks?

LLMs can also be integrated into a SOAP (Subjective, Objective, Assessment, and Plan) framework to create a tool to assist health care workers in daily patient interactions. This tool records and summarizes provider-client interactions, converting audio to structured data, sending next-step reminders, streamlining data collection, and reducing manual data entry to enable more focused patient interactions. Workers can review session details, read transcripts, and add insights anytime. GAI-generated personalized care plans and medication recommendations can combine patient-specific data with evidence-based guidelines.

Moreover, when trained on specialized datasets, GAI and LLMs can aid patients and pharmacists by providing comprehensive information about available medications, possible side effects, and interactions. In health tracking and medication procurement apps, LLMs help patients make informed treatment decisions regarding non-prescription or over-the-counter medicines by offering insights into potential side effects and interactions, allowing them to explore detailed information about their prescribed drugs, or suggesting suitable alternatives.

## Ethical AI

In the transformative realm of digital health care, AI offers notable prospects for enhanced decision-making support and more efficient patient care. It also poses significant ethical challenges, particularly around equity in health care resource access, patient privacy, and data security. Addressing these challenges, in addition to bias prevention, fairness, and reliability, is pivotal. Efforts to establish ethical frameworks and guidelines for this field are ongoing.[107–111]

The lag between AI technological advancements and developing adequate regulatory and ethical understanding frameworks is evident. Achieving AI beneficial to societal needs necessitates transparency, accountability, and most importantly, stakeholder engagement, throughout its development and deployment. Essential voices in discussions shaping the ethical landscape of AI in healthcare range from patient groups and community leaders to national health care authorities.

Concerning privacy and data protection, adherence to regulatory standards, including the GDPR HIPAA, is imperative. These standards require that data, such as health worker and patient information, is anonymized and devoid of direct identifiers, providing a strong layer of protection for individuals. Mechanisms should also facilitate user data deletion requests and accommodate different sensitivities of data through varied anonymization levels.[91,92,112]

Ensuring ML models are reliable and safe also involves scrutinizing performance, understanding biases, and guaranteeing reasonable accuracy for the intended use and target demographics. Strategies such as pursuing representative datasets and implementing bias mitigation strategies are essential to ensure fairness and accuracy in model recommendations. One approach to achieving explicit fairness integration into models and decision algorithms is through multi-objective optimization, which involves balancing multiple competing objectives, such as accuracy and fairness, to find the best possible solution.[92,93,113] Utilizing interpretable models and causal inference in certain use cases can also enhance transparency and accountability in AI healthcare systems.[76,114,115] Facilitating experimentation with various intervention strategies is crucial to achieving fairness. It allows systematic testing of algorithmic-based feature impacts in real-world scenarios, including exploring effects among diverse social groups. Although the system leans on data and models for user support, the ultimate decision-maker remains the human-in-the-loop, with user feedback and validation integral to enhancing recommendation accuracy and fairness.

These ethical concerns, while particularly poignant for health care applications due to their high stakes, apply to any AI system. Applying AI in a health systems framework raises additional ethical concerns around the use of these technologies by stakeholders whose interests may conflict with public health advancement. For example, pharmacies could exploit these technologies to increase their profits by selling antimalarials or antibiotics when not required, contravening patient welfare and public health interest. While AI might exacerbate the danger, these are often preexisting challenges. However, technology also equips us with the means to counter them: transparency and systematic supervision. Digitalization creates data trails and methods to audit them when certain operations could damage public health interests. Regulatory frameworks and incentives must be in place to ensure supervisory bodies have access to these data, to the tools (including our platform) to extract meaning from them, and to a variety of positive and negative incentives to nudge behavior toward our collective interest in improving public health.

The development and deployment of AI in health care needs clear ethical guidelines that stringently address principles such as privacy, reliability, fairness, and user agency, underscored by transparency and accountability. Engaging all stakeholders, especially those typically underrepresented, in constructive discussions will be pivotal in shaping future directions and ethical standards. Regulation and supervision can mitigate the risks to public health arising from the capacities AI can endow. Therefore, it is crucial that we prioritize the development of these guidelines and the engagement of all stakeholders in the process.

## Summary and Conclusions

Pharmacies, health care facilities, and health workers are essential components in the effort to create effective access to quality health care for everyone everywhere. However, in LMICs, fragmented health ecosystems—dominated by informal retailers, the circulation of substandard and falsified medicines, and scarcity of resources—do not allow providers and health care systems to perform to their optimal capabilities. Many digital tools are emerging to contribute to improving this situation. Such technologies include capacity-building tools, tools that provide health care workers with needed supplies (e.g., drug delivery services) or medical resources (e.g., arrangement of test appointments), tools that connect them to patients (e.g., for a remote follow-up) or other physicians (e.g., to get their opinion on test results), and even apps to assist in clinical triage and diagnosis. They bring efficiency, and users generate robust sequential data into the health care delivery model. The data generated by these tools contains rich information on the users' behaviors, preferences, biases, and needs, as well as the evolution of the demand for different supplies.

While collection and organization of information is the key to any data-driven developments, its greatest potential comes from the ability to transform that data into adaptive interventions to support users in a personalized manner to impact patients' health outcomes and lives. Interventions are adaptive in time and to each user's context at a given moment. Adaptive algorithms can also be used for fair and efficient resource allocation.

This article presents an operational RL platform that uses data, predictive modeling, adaptive experimentation, and bandit algorithms to enhance mobile applications for health and health care delivery using personalized content, incentives, reminders, and recommendations, including resource allocation. The platform allows the definition of controlled trials following different experimental designs to establish the effectiveness of the interventions. Experiments have shown that the platform contributes to improved operational efficiency and reduced stock deficiencies by optimizing information and encouraging all actors toward better-informed decisions in inventory management and patient care.

Furthermore, the framework offers different levels of insights about the health system to decision-makers, including early alerts for out-of-norm events and when things are evolving differently than expected. The AI platform can be used to simulate the effects of different strategies or policies through adaptive experimentation. It can thus play a crucial role in guiding evidence-based decision-making to drive the efficiency, effectiveness, and sustainability of health systems. The potential impact of this approach on system efficiency and health outcomes is beneficial in any setting, regardless of the level of resources; however, it is potentially most critical for LMICs, where it can compensate for resource scarcity.

## Notes

[a] For more technical audiences: While API integration for very specific use cases and punctual data sources might be considered in the future, an SDK is our chosen method because it allows for seamless integration with the application, ensuring consistent real-time tracking of in-app activity without impacting app performance.

[b] For more technical audiences: While the use of pre-trained models might become relevant in the future, the approach so far has been to train all models from scratch. The platform is designed to allow for easy and flexible model definition, picking the architecture and data found to be suitable for each use case, digital tool, and context. In the case of RL-based adaptive intervention delivery, described in subsequent sections, the beliefs used to cold start the system can be informed by data from similar interventions and then sequentially updated through interactions with the current environment.

## Disclosure Statement

The authors reported there is no funding associated with the work featured in this article.

## Funding

The work was supported by the Bill & Melinda Gates Foundation [INV- 053824].

## Acknowledgments

This work was supported, in whole or in part, by the Bill & Melinda Gates Foundation INV- 053824. Under the grant conditions of the Foundation, a Creative Commons Attribution 4.0 Generic License has been assigned to the Author Accepted Manuscript version that might arise from this submission.

An earlier version of this paper was presented at the 40th Anniversary Takemi Symposium on "Digital Health: Opportunities and Challenges for Global Health," held on 20-21 October 2023, in Boston, MA, at Harvard University. The Symposium was organized by the Takemi Program in International Health at the Harvard T.H. Chan School of Public Health, with support from the Japan Medical Association and sponsorship from the Japan Pharmaceutical Manufacturers Association, the Takemi Memorial Trust for Research of Seizon and Life Sciences, and the Pharmaceutical Research and Manufacturers of America.

## References

1. World Health Organization. Working together for health: the world health report 2006. Geneva: World Health Organization; 2006.
2. Lippeveld T, Sauerborn R, Bodart C. Design and implementation of health information systems. Geneva, Switzerland: World Health Organization; 2000.
3. World Health Organization Commission on Social Determinants of Health. Closing the gap in a generation: health equity through action on the social determinants of health. final report of the commission on social determinants of health. Geneva: World Health Organization; 2008.
4. Harman S. Global health governance. London: Routledge; 2012.
5. Allemani C, Weir HK, Carreira H, Harewood R, Spika D, Wang XS, Bannon F, Ahn JV, Johnson CJ, Bonaventure A, et al. Global surveillance of cancer survival 1995-2009: analysis of individual data for 25 676 887 patients from 279 population-based registries in 67 countries (concord-2). The Lancet. 2015;385(9972):977–1010. doi:10.1016/S0140-6736(14)62038-9.
6. OECD. Ready for the next crisis? investing in health system resilience. Organisation For Econ Co-Operation And Devel. 2023; doi:10.1787/1e53cf80-en.
7. Abernethy A, Adams L, Barrett M, Bechtel C, Brennan P, Butte A, Faulkner J, Fontaine E, Friedhoff S, Halamka J, et al. The promise of digital health: then, now, and the future. NAM Perspect. 2022;6(22). doi:10.31478/202206e. PMID: 36177208.
8. World Health Organization. Classification of digital health interventions v1.0: a shared language to describe the uses of digital technology for health. Geneve, Switzerland: Technical report, World Health Organization; 2018.

9. Wahl B, Cossy-Gantner A, Germann S, Schwalbe NR. Artificial intelligence (ai) and global health: how can ai contribute to health in resource-poor settings? BMJ Global Health. 2018;3(4):e000798. doi:10.1136/bmjgh-2018-000798.
10. McCarthy M, Birney E. Personalized profiles for disease risk must capture all facets of health. Nature. 2021;597 (7875):175–77. doi:10.1038/d41586-021-02401-0.
11. Subbiah V. The next generation of evidence-based medicine. Nat Med. 2023;29(1):49–58. doi:10.1038/s41591-022-02160-z.
12. Hosny A, Aerts HJ. Artificial intelligence for global health. Science. 2019;366(6468):955–56. doi:10.1126/science.aay5189.
13. Nahum-Shani I, Smith SN, Spring BJ, Collins LM, Witkiewitz K, Tewari A, Murphy SA. Just-in-time adaptive interventions (jitais) in mobile health: key components and design principles for ongoing health behavior support. Ann Behavioral Med. 2018;52 (6):446–62. doi:10.1007/s12160-016-9830-8.
14. Menictas M, Rabbi M, Klasnja P, Murphy S. Artificial intelligence decision-making in mobile health. The Biochemist. 2019;41(5):20–24. doi:10.1042/BIO04105020.
15. Carpenter SM, Menictas M, Nahum-Shani I, Wetter DW, Murphy SA. Developments in mobile health just-in-time adaptive interventions for addiction science. Curr Addict Rep. 2020;7(3):280–90. doi:10.1007/s40429-020-00322-y.
16. Bidargaddi N, Schrader G, Klasnja P, Licinio J, Murphy S. Designing m-health interventions for precision mental health support. Transl Psychiatry. 2020;10 (1). doi:10.1038/s41398-020-00895-2.
17. Coppersmith DD, Dempsey W, Kleiman E, Bentley K, Murphy S, Nock M. Just-in-time adaptive interventions for suicide prevention: promise, challenges, and future directions. Psychiatry. 2021;85(4):317–333. doi:10.1080/00332747.2022.2092828.
18. Hrnjic E, Tomczak N. Machine learning and behavioral economics for personalized choice architecture. 2019. https://arxiv.org/abs/1907.02100.
19. Geng T, Lin X, Nair HS, Hao J, Xiang B, Fan S. Comparison lift: bandit-based experimentation system for online advertising. CoRR. 2020 abs/2009.07899. https://arxiv.org/abs/2009.07899.
20. Liu Y, Li L. A map of bandits for e-commerce. KDD 2021 Workshop on Multi-Armed Bandits and Reinforcement Learning (MARBLE); 2021. https://www.amazon.science/publications/a-map-of-bandits-for-e-commerce.
21. Gimbel S, Kawakyu N, Dau H, Unger JA. A missing link: hiv-/aids-related mHealth interventions for health workers in low- and middle-income countries. Curr HIV/AIDS Rep. 2018;15(6):414–22. doi:10.1007/s11904-018-0416-x.
22. Benski AC, Schmidt NC, Viviano M, Stancanelli G, Soaroby A, Reich MR. Improving the quality of antenatal care using mobile health in Madagascar: five-year cross-sectional study. JMIR mHealth uHealth. 2020;8 (7):e18543. doi:10.2196/18543.
23. Kasy M, Sautmann A. Adaptive treatment assignment in experiments for policy choice. Econo- metrica. 2021. ISSN 1468-0262. 89(1):113–32. doi:10.3982/ECTA17527.
24. Major L, Francis GA, Tsapali M. The effectiveness of technology-supported personalised learning in low- and middle-income countries: a meta-analysis. Br J Educ Technol. 2021. 52(5):1935–64. ISSN 1467-8535. doi:10.1111/bjet.13116.
25. Guitart A, Del Río A, Periáñez Á, Bellhouse L. Midwifery learning and forecasting: predicting content demand with user-generated logs. Artif Intel in Med. 2023;138. doi:10.1016/j.artmed.2023.102511.
26. Bzdok D, Altman N, Krzywinski M. Statistics versus machine learning. Nat Methods. 2018;15(4):233–34. ISSN 1548-7105. doi:10.1038/nmeth.4642.
27. Yu KH, Beam AL, Kohane IS. Artificial intelligence in healthcare. Nat Biomed Eng. 2018;2(10):719–31. doi:10.1038/s41551-018-0305-z.
28. Yu C, Liu J, Nemati S, Yin G. Reinforcement learning in healthcare: a survey. ACM Comput Surv. 2023 Nov;55 (1):1–36. ISSN 0360-0300. doi:10.1145/3477600.
29. Liu S, See K, Ngiam K, Celi L, Sun X, Feng M. Reinforcement learning for clinical decision support in critical care: comprehensive review. J Med Internet Res. 2020;22(7):e18477. doi:10.2196/18477.
30. Liu M, Shen X, Pan W. Deep reinforcement learning for personalized treatment recommendation. Stat Med. 2022 June;41(20):4034–56. doi:10.1002/sim.9491.
31. Du X, King AA, Woods RJ, Pascual M. Evolution-informed forecasting of seasonal influenza a (H3N2). Sci Transl Med. 2017;9(413). doi:10.1126/scitranslmed.aan5325.
32. Althouse BM, Ng YY, Cummings DA. Early prediction of dengue fever outbreaks using machine learning algorithms. Sci Rep. 2018;8(1). doi:10.1038/s41598-018-34107-w.
33. Liu Y, Wang W, Jiang JQ, Zhang M, Wan KL, Hamzah H, Lim C, Tham Y-C, Cheung CY, Tai ES, et al. A deep learning algorithm to detect chronic kidney disease from retinal photographs in community-based populations. Lancet Digit Health. 2020;2(6):e295–302. doi:10.1016/S2589-7500(20)30063-7.
34. Olaniyi BY, Del Río AF, Periáñez Á, Bellhouse L. User Engagement in Mobile Health Applications. Proceedings of the 28th ACM SIGKDD Conference on Knowledge Discovery and Data Mining; Aug 14, 2022 – Aug 18, 2022; Washington DC, U.S. New York, NY, United States: ACM; 2022. p. 4704–4712 doi:10.1145/3534678.35426.
35. Guitart A, Del Río AF, Periáñez Á, Bellhouse L. Midwifery learning and forecasting: predicting content demand with user-generated logs. Artif Intell Med. 2023;138:102511. doi:10.1016/j.artmed.2023.102511.
36. Guitart A, Heydari A, Olaleye E, Ljubicic J, Del Río AF, Periáñez Á, Bellhouse L. A recommendation system to enhance midwives' capacities in low-income countries. NeurIPS Machine Learning in Public Health workshop (MLPH 2021); December 6 – 14, 2021; Virtual-only Conference. 2021. https://arxiv.org/abs/2111.01786.
37. Katsaris I, Vidakis N. Adaptive e-learning systems through learning styles: a review of the literature. Adv Mob Learn Educ Res. 2021;1(2):124–45. doi:10.25082/amler.2021.02.007.
38. Hardeman W, Houghton J, Lane K, Jones A, Naughton F. A systematic review of just-in-time

adaptive interventions (jitais) to promote physical activity. Int J Behav Nutr Phys Act. 2019;16(1):31. doi:10.1186/s12966-019-0792-7.
39. Bitton A, Fifield J, Ratcliffe H, Karlage A, Wang H, Veillard JH, Schwarz D, Hirschhorn LR. Primary healthcare system performance in low-income and middle-income countries: a scoping review of the evidence from 2010 to 2017. BMJ Global Health. 2019 Aug; 4(Suppl 8):e001551. ISSN 2059-7908. doi:10.1136/bmjgh-2019-001551. Publisher: BMJ Specialist Journals Section: Research.
40. mothers2mothers. https://m2m.org/.
41. Lifebank. https://lifebankcares.com/.
42. Ilara Health, Health management information system. https://www.ilarahealth.com/hmis.
43. Global Strategies, Noviguide: Scaling quality healthcare. https://www.globalstrategies.org/projects/noviguide.
44. Maternity Foundation, Safe delivery app. [accessed 2021 Sep 3]. https://www.maternity.dk/safe-delivery-app/.
45. Lund S, Boas IM, Bedesa T, Fekede W, Nielsen HS, Sørensen BL. Association between the safe delivery app and quality of care and perinatal survival in Ethiopia: a randomized clinical trial. JAMA Pediatr. 2016 Aug;170(8):765–71. ISSN 2168-6203. doi:10.1001/jamapediatrics.2016.0687.
46. Thompson S, Mercer MA, Hofstee M, Stover B, Vasconcelos P, Meyanathan S. Connecting mothers to care: effectiveness and scale-up of an mHealth program in timor-leste. J Global Health. 2019 Dec; 9 (2):020428. ISSN 2047-2978. doi:10.7189/jogh.09.020428.
47. Momentum app. https://aidechemist.com/.
48. Aya contigo. https://hola.ayacontigo.org/.
49. Chekkit. https://chekkitapp.com/.
50. Appy saúde. https://appysaude.co.ao/home.
51. Swiperx pharmacy solutions. https://www.swiperx.com/ .
52. Field intelligence. https://field.inc/.
53. Vledder M, Friedman J, Sjöblom M, Brown T, Yadav P. Improving supply chain for essential drugs in low-income countries: results from a large scale randomized experiment in zambia. Health Syst & Reform. 2019;5(2):158–77. doi:10.1080/23288604.2019.1596050. PMID: 31194645.
54. Levine R, Pickett J, Sekhri N, Yadav P. Demand forecasting for essential medical technologies. Am J Law & Med. 2008;34(2–3):225–55. doi:10.1177/009885880803400206.
55. Wright MN, Dankowski T, Ziegler A. Unbiased split variable selection for random survival forests using maximally selected rank statistics. Stat Med. 2017;36 (8):1272–84. doi:10.1002/sim.7212.
56. Fu W, Simonoff JS. Survival trees for left-truncated and right-censored data, with application to time-varying covariate data. Biostatistics. 2016;18(2):352–69. ISSN 1468–4357. doi:10.1093/biostatistics/kxw047.
57. Lee C, Yoon J, van der Schaar M. Dynamic-deephit: a deep learning approach for dynamic survival analysis with competing risks based on longitudinal data. IEEE Trans On Biomed Eng. 2020;67(1):122–33. doi:10.1109/tbme.2019.2909027.
58. Benidis K, Rangapuram SS, Flunkert V, Wang Y, Maddix D, Turkmen C, Gasthaus J, Bohlke-Schneider M, Salinas D, Stella L, et al. Deep learning for time series forecasting: tutorial and literature survey. ACM Comput Surv. 2023 Dec;55(6):1–36. ISSN 0360-0300. doi:10.1145/3533382.
59. Yu HF, Rao N, Dhillon IS. Temporal regularized matrix factorization for high-dimensional time series prediction. Adv Neural Inf Process Syst. 2016;29. Curran Associates, Inc. https://papers.nips.cc/paper_files/paper/2016/hash/85422afb467e9456013a2a51d4dff702-Abstract.html.
60. Seeger M, Rangapuram S, Wang Y, Salinas D, Gasthaus J, Januschowski T, Flunkert V. Approximate bayesian inference in linear state space models for intermittent demand forecasting at scale. 2017 Sep. http://arxiv.org/abs/1709.07638.
61. Salinas D, Bohlke-Schneider M, Callot L, Medico R, Gasthaus J. High-dimensional multivariate forecasting with low-rank gaussian copula processes. Adv Neural Inf Process Syst. 2019;32. Curran Associates, Inc. https://proceedings.neurips.cc/paper_files/paper/2019/hash/0b105cf1504c4e241fcc6d519ea962fb-Abstract.html.
62. Salinas D, Flunkert V, Gasthaus J, Januschowski T. Deepar: probabilistic forecasting with autore- gressive recurrent networks. Int J Forecasting. 2020 Jul; 36 (3):1181–91. ISSN 0169-2070. doi:10.1016/j.ijforecast.2019.07.001.
63. Lim B, Arık SO, Loeff N, Pfister T. Temporal fusion transformers for interpretable multi-horizon time series forecasting. Int J Forecasting. 2021;Oct. 37 (4):1748–64. ISSN 0169-2070. doi:10.1016/j.ijforecast.2021.03.012.
64. Qu Y, Cai H, Ren K, Zhang W, Yu Y, Wen Y, Wang J. Product-based neural networks for user response prediction. 2016 IEEE 16th International Conference on Data Mining (ICDM); 2016; IEEE. p. 1149–54. doi:10.1109/ICDM.2016.0151.
65. Guo H, Tang R, Ye Y, Li Z, He X. Deepfm: a factorization-machine based neural network for ctr prediction. Proceedings of the 26th International Joint Conference on Artificial Intelligence; August 19 – 25, 2017; Melbourne Australia. AAAI Press; 2017. p. 1725–1731.
66. Lian J, Zhou X, Zhang F, Chen Z, Xie X, Sun G. Combining explicit and implicit feature interactions for recommender systems. Proceedings of the 24th ACM SIGKDD International Conference on Knowledge Discovery & Data Mining; 2018; ACM. p. 1754–63. doi:10.1145/3219819.3220023.
67. Lu W, Yu Y, Chang Y, Wang Z, Li C, Yuan B. A dual input-aware factorization machine for ctr prediction. IJCAI'20: Proceedings of the Twenty-Ninth International Joint Conference on Artificial Intelligence; 2020. p. 3139–45. https://www.ijcai.org/proceedings/2020/0434.pdf.
68. O'Connor S, Hanlon P, O'Donnell CA, Garcia S, Glanville J, Mair FS. Understanding factors affecting patient and public engagement and recruitment to digital health interventions: a systematic review of qualitative studies. BMC Med Inf Decis Mak. 2016;16(1):120. ISSN 1472-6947. doi:10.1186/s12911-016-0359-3.
69. Klasnja P, Hekler EB, Shiffman S, Boruvka A, Almirall D, Tewari A, Murphy SA. Microrandomized

trials: an experimental design for developing just-in-time adaptive interventions. Health Phychol. 2015;34 (Suppl):1220–28. ISSN 1930-7810. doi:10.1037/hea0000305. Place: US Publisher: American Psychological Association.

70. Qian T, Walton AE, Collins LM, Klasnja P, Lanza ST, Nahum-Shani I, Rabbi M, Russell MA, Walton MA, Yoo H, et al. The microrandomized trial for developing digital interventions: experimental design and data analysis considerations. Psycholog Met. 2022 Oct;27 (5):874–94. ISSN 1939-1463. doi:10.1037/met0000283.

71. Zhang KW, Janson L, Murphy SA. Statistical inference after adaptive sampling in non-markovian environments. 2022. https://arxiv.org/abs/2202.07098.

72. Burtini G, Loeppky J, Lawrence R. A survey of online experiment design with the stochastic multi-armed bandit. 2015. https://arxiv.org/abs/1510.00757.

73. Yao J, Brunskill E, Pan W, Murphy S, Doshi-Velez F. Power constrained bandits. In: Jung K, Yeung S, Sendak M, Sjoding M, Ranganath R. editors. Proceedings of the 6th machine learning for healthcare conference, volume 149 of proceedings of machine learning research. PMLR; 2021. p. 209–59. https://proceedings.mlr.press/v149/yao21a.html.

74. Dwivedi R, Tian K, Tomkins S, Klasnja P, Murphy S, Shah D. Counterfactual inference for sequential experiments. 2023 Apr. http://arxiv.org/abs/2202.06891 .

75. Xiang D, West R, Wang J, Cui X, Huang J. Multi armed bandit vs. a/b tests in e-commerce - confidence interval and hypothesis test power perspectives. Proceedings of the 28th ACM SIGKDD Conference on Knowledge Discovery and Data Mining, KDD '22; 2022; (NY), NY, USA: Association for Computing Machinery; p. 204–14. ISBN 9781450393850. doi:10.1145/3534678.3539144.

76. Imbens GW, Rubin DB. Causal inference for statistics, social, and biomedical sciences. Cambridge University Press; 2015 Apr. doi:10.1017/cbo9781139025751.

77. Imai K, King G, Nall C. The essential role of pair matching in cluster-randomized experiments, with application to the Mexican universal health insurance evaluation. Stat Sci. 2009 Feb;24(1). doi:10.1214/08-sts274.

78. Bai Y. Optimality of matched-pair designs in randomized controlled trials. Am Econ Rev. 2022 Dec;112 (12):3911–40. doi:10.1257/aer.20201856.

79. Sutton RS, Barto AG. Reinforcement learning: an introduction. Cambridge (MA), USA: A Bradford Book; 2018 Oct. ISBN 978-0-262-03924-6.

80. Lattimore T, Szepesvári C. Bandit algorithms. Cambridge: Cambridge University Press; 2020. ISBN 978-1-108-48682-8. doi:10.1017/9781108571401.

81. Li L, Chu W, Langford J, Schapire RE. A contextual-bandit approach to personalized news article recommendation. Proceedings of the 19th international conference on World wide web, WWW '10; 2010 Apr; (NY), NY, USA: Association for Computing Machinery; p. 661–70. ISBN 978-1- 60558-799-8. doi:10.1145/1772690.1772758.

82. Chu W, Li L, Reyzin L, Schapire R. Contextual bandits with linear payoff functions. Proceedings of the Fourteenth International Conference on Artificial Intelligence and Statistics; 2011 June; JMLR Workshop and Conference Proceedings; p. 208–14. ISSN: 1938-7228. https://proceedings.mlr.press/v15/chu11a.html.

83. Agrawal S, Goyal N. Thompson sampling for contextual bandits with linear payoffs. Proceedings of the 30th International Conference on Machine Learning; 2013 May; PMLR; p. 127–35. ISSN: 1938-7228. https://proceedings.mlr.press/v28/agrawal13.html.

84. Riquelme C, Tucker G, Snoek J. Deep bayesian bandits showdown: an empirical comparison of bayesian deep networks for thompson sampling. International Conference on Learning Representations; 2018. https://openreview.net/forum?id=SyYe6k-CW.

85. Zhou D, Li L, Gu Q. Neural contextual bandits with ucb-based exploration. Proceedings of the 37th International Conference on Machine Learning; 2020 Nov; PMLR; p. 11492–502. ISSN: 2640-3498. https://proceedings.mlr.press/v119/zhou20a.html.

86. Xu P, Wen Z, Zhao H, Gu Q. Neural contextual bandits with deep representation and shallow exploration. International Conference on Learning Representations; 2022. https://openreview.net/forum?id=xnYACQquaGV.

87. Nabati O, Zahavy T, Mannor S. Online limited memory neural-linear bandits with likelihood matching. Proceedings of the 38th International Conference on Machine Learning; 2021 Jul; PMLR; p. 7905–15. ISSN: 2640-3498. https://proceedings.mlr.press/v139/nabati21a.html.

88. Fedus W, Ramachandran P, Agarwal R, Bengio Y, Larochelle H, Rowland M, Dabney W. Revisiting fundamentals of experience replay. Proceedings of the 37th International Conference on Machine Learning; 2020 Nov; PMLR; p. 3061–71. ISSN: 2640-3498. https://proceedings.mlr.press/v119/fedus20a.html.

89. Duran-Martin G, Kara A, Murphy K. Efficient online bayesian inference for neural bandits. Proceedings of The 25th International Conference on Artificial Intelligence and Statistics; 2022 May; PMLR; p. 6002–21. ISSN: 2640-3498. https://proceedings.mlr.press/v151/duran-martin22a.html.

90. Daum FE. Extended kalman filters. In: Encyclopedia of systems and control. Springer London; 2015. p. 411–13. doi:10.1007/978-1-4471-5058-9_62.

91. Drugan MM, Nowe A. Designing multi-objective multi-armed bandits algorithms: a study. The 2013 International Joint Conference on Neural Networks (IJCNN); 2013 Aug; p. 1–8. ISSN: 2161-4407. doi:10.1109/IJCNN.2013.6707036.

92. Turgay E, Oner D, Tekin C. Multi-objective contextual bandit problem with similarity information. Proceedings of the Twenty-First International Conference on Artificial Intelligence and Statistics; 2018 Mar; PMLR; p. 1673–81. ISSN: 2640-3498. https://proceedings.mlr.press/v84/turgay18a.html.

93. Lu S, Wang G, Hu Y, Zhang L. Multi-objective generalized linear bandits. Proceedings of the Twenty-Eighth International Joint Conference on Artificial Intelligence, IJCAI-19; 2019 Jul; Macao, China:

IJCAI'19; p. 3080–86. ISBN 978-0-9992411-4-1. doi:10.24963/ijcai.2019/427.

94. Lopez R, Dhillon IS, Jordan MI. Learning from extreme bandit feedback. Proc AAAI Conf On Artif Intel. 2021 May;35(10):8732–40. ISSN 2374-3468. doi:10.1609/aaai.v35i10.17058.
95. Bubeck S, Munos R, Stoltz G. Pure exploration in finitely-armed and continuous-armed bandits. Theor Comput Sci. 2011 Apr;412(19):1832–52. ISSN 0304-3975. doi:10.1016/j.tcs.2010.12.059.
96. Whittle P. Restless bandits: activity allocation in a changing world. J Appl Probab. 1988;25:287–98. ISSN 0021-9002. doi:10.2307/3214163.
97. Qian Y, Zhang C, Krishnamachari B, Tambe M. Restless poachers: handling exploration- exploitation tradeoffs in security domains. Proceedings of the 2016 International Conference on Autonomous Agents & Multiagent Systems, AAMAS '16; 2016 May; Richland (SC): Inter- national Foundation for Autonomous Agents and Multiagent Systems. ISBN 978-1-4503-4239-1 123–31. https://dl.acm.org/doi/10.5555/2936924.2936946 .
98. Bagheri S, Scaglione A. The restless multi-armed bandit formulation of the cognitive compressive sensing problem. IEEE Trans On Signal Process. 2015 Mar;63(5):1183–98. ISSN 1941-0476. doi:10.1109/TSP.2015.2389620.
99. Mate A, Madaan L, Taneja A, Madhiwalla N, Verma S, Singh G, Hegde A, Varakantham P, Tambe M. Field study in deploying restless multi-armed bandits: assisting non-profits in improving maternal and child health. Proc AAAI Conf On Artif Intel. 2022 June;36(11):12017–25. ISSN 2374-3468. doi:10.1609/aaai.v36i11.21460.
100. Biswas A, Aggarwal G, Varakantham P, Tambe M. Learn to intervene: an adaptive learning policy for restless bandits in application to preventive healthcare. In Zhou Z. editor. Proceedings of the Thirtieth International Joint Conference on Artificial Intelligence, IJCAI-21; 2021 8; International Joint Conferences on Artificial Intelligence Organization,; p. 4039–46. doi:10.24963/ijcai.2021/556.
101. Nishtala S, Madaan L, Mate A, Kamarthi H, Grama A, Thakkar D, Narayanan D, Chaudhary S, Madhi-Walla N, Padmanabhan R, et al. Selective intervention planning using restless multi-armed bandits to improve maternal and child health outcomes. 2021 Oct. http://arxiv.org/abs/2103.09052.
102. Lee E, Lavieri MS, and Volk M. Optimal screening for hepatocellular carcinoma: a restless bandit model. Manuf & Service Operations Manag. 2019 Jan;21(1):198–212. ISSN 1523-4614. doi:10.1287/msom.2017.0697.
103. Bhattacharya B. Restless bandits visiting villages: a preliminary study on distributing public health services. Proceedings of the 1st ACM SIGCAS Conference on Computing and Sustainable Societies, COMPASS '18; 2018 June; (NY), NY, USA: Association for Computing Machinery. 1–8. ISBN 978-1-4503-5816-3. doi:10.1145/3209811.3209865.
104. Baek J, Boutilier JJ, Farias VF, Jonasson JO, Yoeli E. Policy optimization for personalized interven- tions in behavioral health. 2023 Mar. http://arxiv.org/abs/2303.12206. arXiv:2303.12206[cs].
105. Killian JA, Perrault A, Tambe M. Beyond"to act or not to act": fast lagrangian approaches to general multi-action restless bandits. Proceedings of the 20th International Conference on Autonomous Agents and MultiAgent Systems; May 3 – 7, 2021; Virtual Event United Kingdom; 2021. p. 710–18.
106. Ou HC, Siebenbrunner C, Killian J, Brooks MB, Kempe D, Vorobeychik Y, Tambe M. Networked restless multi-armed bandits for mobile interventions. 2022 Jan. http://arxiv.org/abs/2201.12408. arXiv:2201.12408[cs].
107. Raji ID, Smart A, White RN, Mitchell M, Gebru T, Hutchinson B, Smith-Loud J, Theron D, Barnes P. Closing the ai accountability gap: defining an end-to-end framework for internal algorithmic auditing. Proceedings of the 2020 Conference on Fairness, Accountability, and Transparency, FAT* '20; 2020. (NY), NY, USA: Association for Computing Machinery. p. 33–44. ISBN 9781450369367. doi:10.1145/3351095.3372873.
108. Prabhakaran V, Mitchell M, Gebru T, Gabriel I. A human rights-based approach to responsible ai. 2022. https://arxiv.org/abs/2210.02667.
109. Dubber MD, Pasquale F, Das S, editors. The oxford handbook of ethics of ai. Oxford University Press; 2020 Jul. doi:10.1093/oxfordhb/9780190067397.001.0001.
110. Jobin A, Ienca M, Vayena E. The global landscape of ai ethics guidelines. Nat Mach Intel. 2019 Sep;1(9):389–99. doi:10.1038/s42256-019-0088-2.
111. Chen IY, Pierson E, Rose S, Joshi S, Ferryman K, Ghassemi M. Ethical machine learning in healthcare. Annu Rev Biomed Data Sci. 2021;4(1):123–44. doi:10.1146/annurev-biodatasci-092820-114757.
112. Sweeney L. K-anonymity: a model for protecting privacy. Int J Uncertain Fuzziness Knowl-Based Syst. 2002 Oct;10(5):557–70. ISSN 0218-4885. doi:10.1142/S0218488502001648.
113. Mhasawade V, Zhao Y, Chunara R. Machine learning and algorithmic fairness in public and population health. Nat Mach Intel. 2021 Jul;3(8):659–66. doi:10.1038/s42256-021-00373-4.
114. Glocker B, Musolesi M, Richens J, Uhler C. Causality in digital medicine. Nat Commun. 2021;12(1):Article no 5471. doi:10.1038/s41467-021-25743-9.
115. Rudin C. Stop explaining black box machine learning models for high stakes decisions and use interpretable models instead. Nat Mach Intell. 2019 May;1(5):206–15. doi:10.1038/s42256-019-0048-x.